# Genetic Algorithm enhanced by Deep Reinforcement Learning in parent selection mechanism and mutation : Minimizing makespan in permutation flow shop scheduling problems

M. Irmouli, N. Benazzoug, A.D. Adimi, F.Z. Rezkellah,I. Hamzaoui, T. Hamitouche, M. Bessedik, F. Si Tayeb
Ecole Nationale Supérieure d'Informatique, Algiers, Algeria

**Abstract**

This paper introduces a reinforcement learning (RL) approach to address the challenges associated with configuring and optimizing genetic algorithms (GAs) for solving difficult combinatorial or non-linear problems. The proposed RL+GA method was specifically tested on the flow shop scheduling problem (FSP). The hybrid algorithm incorporates neural networks (NN) and uses the off-policy method Q-learning or the on-policy method Sarsa(0) to control two key genetic algorithm (GA) operators: parent selection mechanism and mutation. At each generation, the RL agent's action is determining the selection method, the probability of the parent selection and the probability of the offspring mutation. This allows the RL agent to dynamically adjust the selection and mutation based on its learned policy. The results of the study highlight the effectiveness of the RL+GA approach in improving the performance of the primitive GA. They also demonstrate its ability to learn and adapt from population diversity and solution improvements over time. This adaptability leads to improved scheduling solutions compared to static parameter configurations while maintaining population diversity throughout the evolutionary process.

## 1. INTRODUCTION

NP-hard problems represent a class of challenging computational problems that encompass a wide range of real-world scenarios. These problems require finding the best possible solutions within a reasonable time frame but lack efficient algorithms for finding optimal solutions within a reasonable time frame [1].

Metaheuristic algorithms came to solve NP-hard problems. They are designed to efficiently explore large solution spaces and incorporate randomness and local search strategies to improve the quality of solutions, offer promising approaches to tackle NP-hard problems by providing heuristic solutions that approximate the optimal or near-optimal solutions that can't be reached using traditional optimization techniques [1].

Moving from our exploration of NP-difficult problems, we now redirect our attention to a specific problem domain : the Flow Shop Scheduling Problem (FSP). It is a well-known classic combinatorial optimization problem, in the permutation FSSP (PFSP), all jobs must enter the machines in the same order and the goal is to find a job permutation that minimizes a specific performance measure, usually makespan or total flowtime [2]. The makespan is the time it takes for all jobs to be processed and completed. PFSP is widely recognized as a computationally complex problem with various practical applications in industries such as manufacturing, logistics, and production planning. As PFSP falls under the category of NP-difficult problems, metaheuristic algorithms have been turned to in order to address its complexities such as Greedy randomized constructive heuristic and Nested Exploration Genetic Algorithm - VNS approach [3].

However, despite effectiveness of metaheuristics for solving FSP and other combinatorial optimization problems, they have inherent limitations that impact their performance and efficiency. including the risk of getting trapped in local optima, high computational requirements that makes them impractical for large-scale problems, also, their sensitivity to parameter settings remains a prominent issue.

From the wide range of metaheuristic algorithms, Genetic Algorithms (GAs) stand out as a well-known and prominent approach. These algorithms specifically operate by simulating the evolution process, where a population of candidate solutions evolves over generations to find an approximation of an optimal solution [4]. GAs often require configuration, operators must be set, and parameters decided including population size, mutation rate, crossover rate, and selection pressure, and so on [5]. Addressing the problem of parameter calibration in GAs is crucial for their successful application. Researchers have explored various methods, including : fuzzy logic [6], design of experiments (DOE) [7], full factorial and response surface methodology experimental designs [8] and

Relevance Estimation And Value Calibration of GA parameters (RE-VAC) [9]. There are also those who head for ML techniques, specifically RL techniques. We mention : RL-GA approach using Q-learning on the traveling salesman problem (TSP) [10] which selects the class of individuals for the mating pool (Fit or Unfit) and the particular crossover or mutation operator. SCGA that addresses the problems of RL-GA and implements the on-policy method Sarsa(0) [11] that can perform without training first. On the other hand, another RL approach was proposed for a GA used for solving a Capacitated Vehicle Routing Problem(CVRP) [12] where the agent selects probabilities of crossover and mutation. In fact RL is widely used for controlling Parameters of Evolutionary Algorithms such as GAs [13].

We can see that the integration of RL with GAs offers a promising avenue for enhancing the performance and effectiveness of GAs in solving optimization problems. The adaptive nature of RL can be leveraged to dynamically optimize the parameters and operators of GAs, leading to improved performance and robustness. RL can learn from experience and adapt its actions based on the feedback received from the environment, which in the case of GA involves adjusting parameters and genetic operators.

RL and GA also have many similarities such as their use of a probabilistic search mechanism to optimize solutions: the evolution of generations in GA can be considered as a learning process, thus GA operators fulfill a role similar to actions in RL.

For a better solution to overcome the limitations of GA's parameters selection, we propose an RL agent that dynamically adjusts the GA's parent selection mechanism and mutation to optimize the scheduling of tasks in the flow shop scenario with the help of NN to select appropriate action for each state. Neural Networks (NNs) are known for their ability to capture complex patterns and make informed decisions [14] . While traditional deciding selection methods in Genetic Algorithms (GA) often rely on simple rules or heuristics, which may not fully exploit the richness of the problem space, by employing NNs, we can leverage their capacity to learn from data and discover intricate relationships between candidate solutions and their corresponding fitness values. We implement two types of agents : an Offline agent uses Deep Q-Learning (DQN) to control over the GA, and an online agent learns its policy by adjusting the NN weights for each generation of GA.

The next section describes the problem formulation, then, The section 3 describes the offline and online version of the algorithm, its architecture, behavior and the structure used for solutions representation with the details of each component. Followed by the experiments section. Finally, the last section that concludes this paper.

## 2. PROBLEM FORMULATION

### 2.1. PFSP

The flow shop scheduling problem determines an optimum sequence of n jobs to be processed on m machines in the same order i.e. every job must be processed on machines *1,2,…,m* in this same order. The flowshop scheduling problem is a production problem where a set of *n* jobs have to be processed with identical flow patterns on *m* machines. When the sequence of jobs processing on all machines is the same we have the permutation flowshop sequencing production environment. We study the flow-shop problems considering the following assumptions [15]:

- The operation processing times on the machines are known, fixed and some of them may be zero when a job is not processed on a machine.
- Set-up times are included in the processing times and they are independent of the job position in the sequence of jobs.
- Each machine can handle only one job at a time.
- Each job is continuously processed on M available machines in the same technological order.

| Machines | Job order | | | | |
|---|---|---|---|---|---|
| Machine 1 | J1 | J2 | J3 | | |
| Machine 2 | | J1 | J2 | J3 | |
| Machine 3 | | | | J1 | J2 | J3 |

Figure 01. Illustration of a flowshop scheduling problem with three jobs and three machines

The total processing time, also known as makespan, refers to the overall duration required to complete all the jobs in a given scheduling problem. It represents a measure of the efficiency and performance of a solution. In the context of scheduling and optimization problems, minimizing the makespan is a common objective.

## 2.2. Genetic Algorithm

Genetic Algorithms (GAs) are randomized search algorithms classified under evolutionary algorithms [16]. They emulate the natural selection process, favoring the survival of the fittest. In a GA, each generation comprises a population of individuals or chromosomes, representing potential solutions within the search space. Through a fitness-based process, individuals are chosen from the population at each iteration, and genetic operators such as crossover and mutation are applied to generate the subsequent population. This iterative approach helps refine and evolve the solutions towards optimal outcomes.

The process of GA consists of the following primary components:

1. **Population Initialization:** In the context of flowshop scheduling problem, a population contains a diverse set of arrays. Each array represents a unique sequence of jobs, indicating the order in which they will be processed. During the initialization phase, the arrays undergo refinement to ensure the sequences are valid and optimal. This involves adjusting the order of jobs within the arrays by considering the machine constraints. By iteratively generating and refining the population, we aim to explore a wide range of potential solutions to the flowshop problem and increase the likelihood of finding optimal or near-optimal solutions.

2. **Fitness Function:** The fitness score quantifies the quality of an individual within the population and plays a vital role in selecting chromosomes for reproduction. In the context of our study, the fitness metric we employ is the makespan.

3. **Selection:** The selection phase aims to identify individuals within each generation that exhibit greater promise and have a higher likelihood of generating improved solutions. This phase involves a fitness-based process, which can be executed through various methods such as random selection, roulette wheel selection, tournament selection, or the use of an elitist function that consistently selects the best individuals.

4. **Reproduction:** The reproduction aims to generate new offspring by combining genetic information from parent individuals. There are two main genetic reproduction operators that are:
    a. **Crossover:** also known as recombination. It is a genetic operator that involves combining genetic information from two parent individuals in order to create offspring. It emulates the biological process of reproduction. During crossover, specific segments of genetic material (genes or alleles) are exchanged between parents,

resulting in offspring that inherit traits from both parents. Crossover promotes the exploration of new solution spaces, enabling the offspring to potentially possess favorable traits from both parents.

  b. **Mutation:** Mutation is a genetic operator that introduces random changes in the genetic material of an individual within a population. It introduces small random modification to one or more genes or alleles of an individual, leading to a slightly different genetic makeup. This random perturbation allows for exploration of new regions in the search space, potentially discovering better solutions that would not be reachable through other operators alone.

5. **Population update:** The offspring, along with some individuals from the current population are used to form the new population for the next generation.

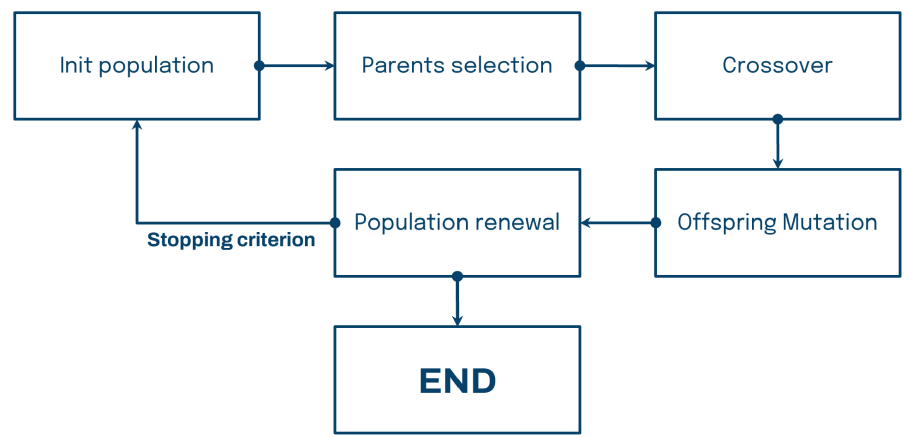

Figure 02. Descriptive Schema of the genetic algorithm

### 2.3. Deep Reinforcement Learning

Reinforcement Learning (RL) is a computational approach to automating goal-directed learning and decision-making [17]. In RL, agents interact with the environment, receiving rewards based on the actions they choose. The primary objective of an RL agent is to learn a policy that maximizes the cumulative rewards obtained over a sequence of actions. Through this learning process, RL agents strive to achieve optimal behavior that leads to the highest possible rewards.

One popular RL algorithm is Q-learning, which aims to learn an optimal action-value function called Q-function. Q-learning utilizes the temporal difference learning paradigm to update Q-values based on the maximum expected future reward [18],

It uses Bellman equation [19]:

$$Q(S_t, a_t) = (1 - \alpha)Q(S_t, a_t) + \alpha\left[r_{t+1} + \gamma maxQ(S_{t+1}, a_t)\right] \quad (1)$$

Action selection follows an epsilon-greedy policy, which randomly chooses an action with a probability of epsilon ($\epsilon$), or selects the action with the highest Q-value with a probability of 1-$\epsilon$. This policy balances exploration and exploitation in decision-making.

Another notable advancement in RL is the Deep Q-Network (DQN), which combines Q-learning with deep neural networks to handle complex and high-dimensional state spaces [20]. DQN can effectively approximate the action-value function, enabling efficient decision-making in challenging environments, thus, DQN is a prominent approach in the field of deep reinforcement learning.

In addition to Q-learning, Sarsa(0) is another RL algorithm commonly used for online learning [21]. Sarsa(0) is an on-policy algorithm that learns directly from interaction with the environment. It stands for *State-Action-Reward-State-Action*, where the agent learns the value of taking an action in a given state and continues the learning process during exploration. Unlike Q-learning, Sarsa(0) considers the next action based on the policy being learned and updates its Q-values accordingly by approximating the function described in **equation (1)**.

## 3. APPROACH DESCRIPTION

### 3.1. General description

The integration of reinforcement learning (RL) with genetic algorithms (GA) introduces a dynamic and adaptive approach to optimize the performance of the GA. The RL agent interacts with the GA environment, leveraging its decision-making capabilities to influence the evolution of the GA. Starting from an initial population, the RL agent explores the search space, which consists of populations generated by the GA. Each population represents a state in the RL agent's learning process.

The RL agent's actions revolve around manipulating two key operators of the GA: **selection** and **mutation**. These actions impact the composition and characteristics of the next generation.

 The RL agent receives rewards that guide its learning process. The reward of each episode is computed through two comparisons : **comparing** the fitness of the child individuals with the parents and **comparing** the best individual in the old and new generation. The RL agent then employs a neural network (NN) model to learn a selection policy for the search operators. The NN model is adjusted and refined based on the learning policy during offline training or in real-time applications for online strategy cases.

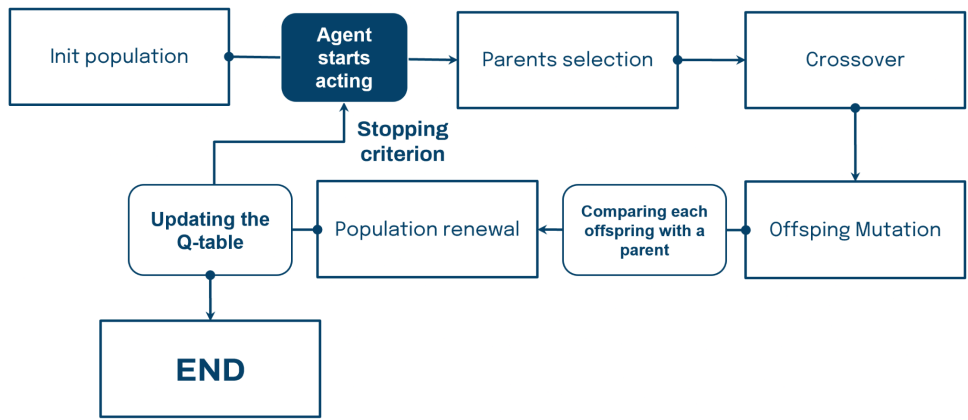

**Figure 03. Descriptive schema of our proposed method**

### 3.2. State representation

To adequately represent the current state of the population in the decision space for the reinforcement learning agent, encoding should capture information regarding the distribution of solutions and their varying fitness values.

A population is represented by a state of two main features:
- Population average fitness.
- The diversity of fitness distribution in the population, calculated using entropy measure [11]:

$$p_m = f(x_m) / \sum_{m=1}^{M} f(x_m) \qquad (2)$$

$$H = \sum_{m=1}^{M} p_m \log_2 \frac{1}{p_m} \qquad (3)$$

Where $M$ is the size of the population and $f(x_m)$ is the fitness of an individual $x_m$.

### 3.3. Actions Definition

The agent can perform a defined set of actions. An action is the tuple $(s, p_s, p_m)$, where $s$ is the parent selection mechanism, $p_s$ and $p_m$ denote the selection and the mutation rate respectively.

There are 3 selection methods, it can be either "Elitism", "Roulette" or "rank", in addition, the ranges of each of $p_s$ and $p_m$ rate from 0 to 1 discretized separately into 3 intervals, and therefore, create $3^3$ = 27 possible ($s$, $p_s$, $p_m$) combinations, meaning the agent chooses one out of 27 possible actions.

### 3.3. Rewards Definition

The reward is determined by the improvement in child individuals in addition to the improvement in the best individual of the population. After the mutation of the children, for each two child individuals generated from a pair of parents we calculate mutation reward as :

**children_reward** = f(child1) + f(child2) - (f(parent1) + f(parent2))    **(4)**

Then, after population renewal, the reward of selection is:

**election_reward** = f(new_best_individual) - f(best_individual)    **(5)**

Where best_individual is the sequence with the minimal fitness in the current population and new_best_individual is the sequence with the minimal fitness in the new generation.

The total reward will be the sum of all children's rewards of the new generation and the difference of fitness between the old and the new best fitness. The reward obtained in the genetic algorithms (GA) can vary between positive and negative based on the fitness improvement achieved in each constructed generation.

The agent's objective is to maximize the reward, thereby incentivizing the selection of operators that generate superior offspring. By following this approach, the GA focuses its exploration on research spaces populated with individuals of high fitness. This mechanism promotes the continuous improvement of the population and directs the evolutionary process towards more optimal solutions.

### 3.4. NN Model Design

The NN serves as a valuable tool for the action selection, allowing the agent to make informed decisions based on its observations and the current state of the environment.

Through the training process, the NN learns to map input information to the corresponding action probabilities. One crucial aspect of the NN architecture is the presence of a softmax activation function in its last layer. The softmax function converts the output values of the NN into a probability distribution over the available actions. This distribution enables the agent to select actions based on their likelihood of achieving favorable outcomes :

$$\pi(S_t, a_t) = \frac{e^{\beta Q(S_t, a_t)}}{\sum_{a_t'=1}^{N} e^{\beta Q(S_t, a_t')}} \qquad (6)$$

The architecture of NN significantly impacts agent performance. Design choices like network depth, width, activation functions, regularization techniques, and specialized layers influence the agent's learning and decision-making abilities.

These aspects must be taken into consideration to enable effective learning, informed decision-making, and improved performance across diverse tasks and environments.

- <u>Learning Algorithm</u>
  <u>Offline approach:</u> uses Q-learning method and needs two phases :
    - The training phase: a neural network is trained to predict the Q-values $Q(S_t, a_t; \theta)$ that are used to select an action for a given state $S_t$ at step t, where θ are weights of the NN.
    - The second phase (online phase): Once the learning is finished, the weights of the NN are frozen and since they are not updated in this phase, we do not maintain computing rewards. As a new state is observed, it is passed to the NN and the Q-values are calculated and then a new action is chosen according to the greedy policy by the agent to be performed by the GA to get a new population and hence a new state.
  
  <u>Online approach:</u> the agent interacts directly with the environment receiving input data and generating output decisions. At each time step, the agent observes the current state, selects an action based on its policy, and interacts with the environment. The agent receives a reward, as previously defined, which reflects the immediate outcome of its action. The agent utilizes Sarsa(0) and updates its NN policy accordingly. Through this dynamic interaction, the agent continuously learns, adapts, and adjusts its behavior, incorporating new experiences to enhance its ability to make informed decisions. This iterative learning process empowers the agent to improve its performance over time by leveraging the feedback received from the environment.

Offline agents benefit from extensive training on a large dataset, leading to improved performance and accurate decision-making. They are not bound by real-time constraints, allowing for thorough analysis. On the other hand, online agents adapt to dynamic environments, updating their neural network in real-time. They actively explore the environment, continuously gathering new data to uncover hidden opportunities. This adaptability enables them to exploit evolving patterns and make informed decisions beyond offline training.

Both agents have their benefits, the choice between offline and online approaches depends on the specific requirements of the problem at hand, including the availability of historical data, real-time constraints, and the need for dynamic exploration.

### 3.5. Genetic operators

In our method, it is important to note that the agent doesn't influence genetic operators, it only chooses actions regarding parents selection and mutation probability. Therefore, for the rest of the components of our GA, we consider the work of **Murata & Al** [22] to choose our genetic operators.

It was found that the best crossover operator for PFSP (Permutation Flowshop Scheduling Problems) is the **two-point crossover version I,** its mechanism is shown in the figure 3. Not only that, they also mention that the best mutation operator for PFSP is a shift mutation, which means a **random insertion,** as shown in figure 4.

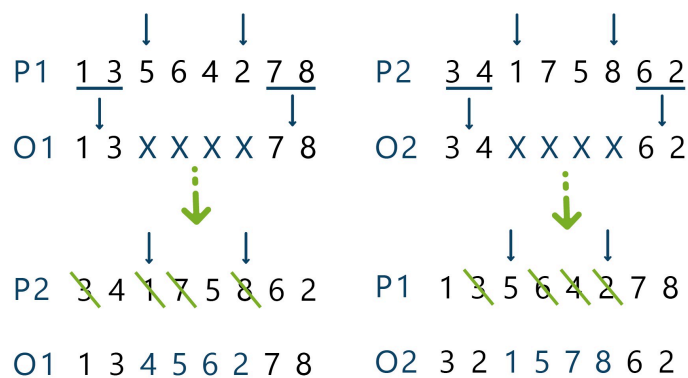

**Figure 04. Modified two-point crossover operator**

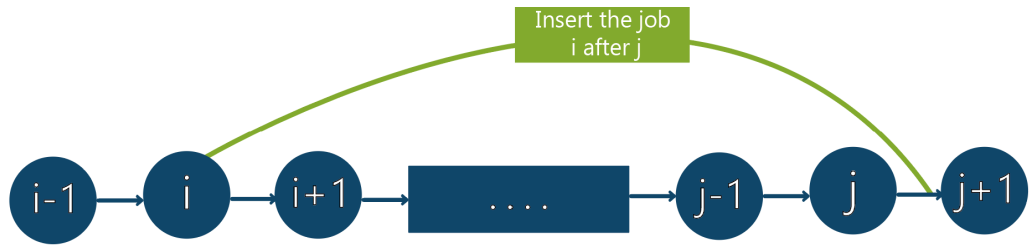

**Figure 05. Shift operator**

## 4. COMPUTATIONAL RESULTS

In this study, we implemented a standard GA and our new hybrid algorithm (offline and online), we are going to present a brief overview of the testing scenario conducted along with discussion of the obtained results in terms of solutions qualities, time execution, and comparison with the standard GA, NEH [23], Greedy NEH [24], CDS [25], VNS [26], Simulated Annealing [27], Tabu Search [28], Stochastic Hill Climbing, NEGA_VNS [29].

The machine used for testing is characterized by :

- **Processor:** Intel(R) Core(TM) i7-6600U CPU @ 2.60GHz   2.81 GHz.
- **RAM:** 16.0 GB (15.9 GB usable).
- **System:** 64-bit operating system, x64-based processor

As for the both online and offline approach, it was executed on the previous machine on CPU and executed on GPU (2xGPU of T4). This choice was made because NN performs more efficiently on the GPU, leading to improved computational performance and faster processing times. The training of the offline agent was also done on the GPU.

Several instances of taillard benchmarks were solved [30], particularly the instances 1 and 7 were selected of the benchmarks: Taillard 20_5 - Taillard 50_10 - Taillard 100_10.

### 4.1. Parameters calibration

Parameter calibration is a critical step in optimizing the performance of our hybrid method. By calibrating RL and GA parameters through a greedy search approach, we aim to achieve optimal performance in the problem domain.

Two groups of parameters intervene:
- **RL parameters** (alpha, gamma, and epsilon), controlling the agent's learning rate, discount factor, and exploration-exploitation trade-off.

- GA parameters (initial population size, number of episodes, and iterations), impacting search space exploration and exploitation.

To handle limited resources and numerous parameters, we employed a greedy search approach. From this search, **we selected alpha = 0.1, gamma = 0.9, and epsilon = 0.5** as optimal RL parameter values.

The greedy search also revealed insights into GA parameters: larger population sizes led to better exploration and improved solution quality, albeit with increased computational time. Smaller population sizes resulted in faster convergence but potentially compromised solution quality. A general rule was derived, suggesting larger population sizes for larger problem instances.

Increasing the number of episodes and iterations enhanced exploration, learning, and convergence. However, resource constraints should be considered. The chosen parameters strike a balance between performance and computational efficiency, ensuring feasible and efficient parameter values.

### Training the agent

For **training** the offline agent, Table 01 displays the parameters associated with the GA, while Table 02 showcases the parameters employed for the RL part of the algorithm. It is important to note that the initial population is generated ***randomly***.

| Benchmark | Number of episodes | Number of iterations | Population size |
|---|---|---|---|
| 20_5 | 50 | 100 | 50 |
| 50_10 | 100 | 200 | 100 |
| 100_10 | 200 | 300 | 200 |

Table 01. Parameters associated with the GA

| parameter | Alpha (α) | Gamma (γ) | Epsilon (ε) |
|---|---|---|---|
| value | 0.1 | 0.9 | 0.5 |

Table 02. parameters employed for the RL part of the algorithm

### Testing the agent

The parameters of the **test** environment are set according to the result of previous parameters calibration, and taking into account the available resources. The parameters related to RL agents still follow those of Table 02, while the parameters related to GA of our approach and of standard GA are found in **Table 03**.

|        | Number of episodes | | Number of iterations | | | Population size | | |
|--------|---------------------|------------------|---------------------|------------------|-----|---------------------|------------------|-----|
|        | CPU DeepRL-GA | GPU DeepRL-GA | CPU DeepRL-GA | GPU DeepRL-GA | GA | CPU DeepRL-GA | GPU DeepRL-GA | GA |
| **20_5**   | 3 | 5 | 50  | 60  | 50  | 30  | 40  | 30  |
| **50_10**  | 5 | 8 | 75  | 85  | 100 | 100 | 120 | 100 |
| **100_10** | 8 | 8 | 100 | 120 | 200 | 120 | 120 | 200 |

Table 03. parameters related to GA of our approach and of standard GA

Additionally, **Table 04** presents the fixed parameters of the standard GA, which are predicted by the agent.

| parameters | Parent selection method | Parent selection rate | Mutation rate |
|------------|------------------------|----------------------|---------------|
| value      | Roulette               | 0.5                  | 0.5           |

Table 04. Fixed parameters of the standard GA

To ensure efficient execution and leverage the knowledge gained from training the offline agent, we have opted for a reduced number of episodes and iterations in the testing phase.

The offline agent has already learned from extensive training, and by using fewer episodes and iterations for the online agent, we aim to strike a balance between performance and computational cost. While increasing these parameters could potentially lead to improved results, it would come at the expense of longer execution times. Thus, our focus is to achieve a reasonable trade-off between efficiency and performance in the online agent's execution.

### 4.2. Performance study
#### 4.2.a. Solution quality

Different results are represented in **Table 05**. The first column on **Table 05** is the sequence ID, each column represents a different algorithm, the second column is our offline agent on CPU, the third is the latter on GPU, the two following columns are reserved for the online version of the algorithm.

As can be seen from **Table 06**, both offline and online show better performances than the standard GA.

| Instance | Published Optimal | Off-CPU DeepRL-GA | Off-GPU DeepRL-GA | On-CPU DeepRL-GA | On-GPU DeepRL-GA | GA |
|---|---|---|---|---|---|---|
| **B_20_5_1** | **1232 *** | 1297 | 1288 | 1282 | 1255 | 1338 |
| **B_20_5_7** | 1226 | 1270 | **1222 *** | 1251 | **1214 *** | 1302 |
| **B_50_10_1** | 2907 | 3091 | 2907 | 3045 | **2902 *** | 3444 |
| **B_50_10_7** | 3062 | 3136 | **3057 *** | 3141 | **3057 *** | 3504 |
| **B_100_10_1** | 5759 | 5798 | **5750 *** | **5748 *** | **5731 *** | 6095 |
| **B_100_10_7** | **5523 *** | 5619 | 5574 | 5573 | 5528 | 60902 |

Table 05. Comparaison through Fitness of different versions of our method with GA

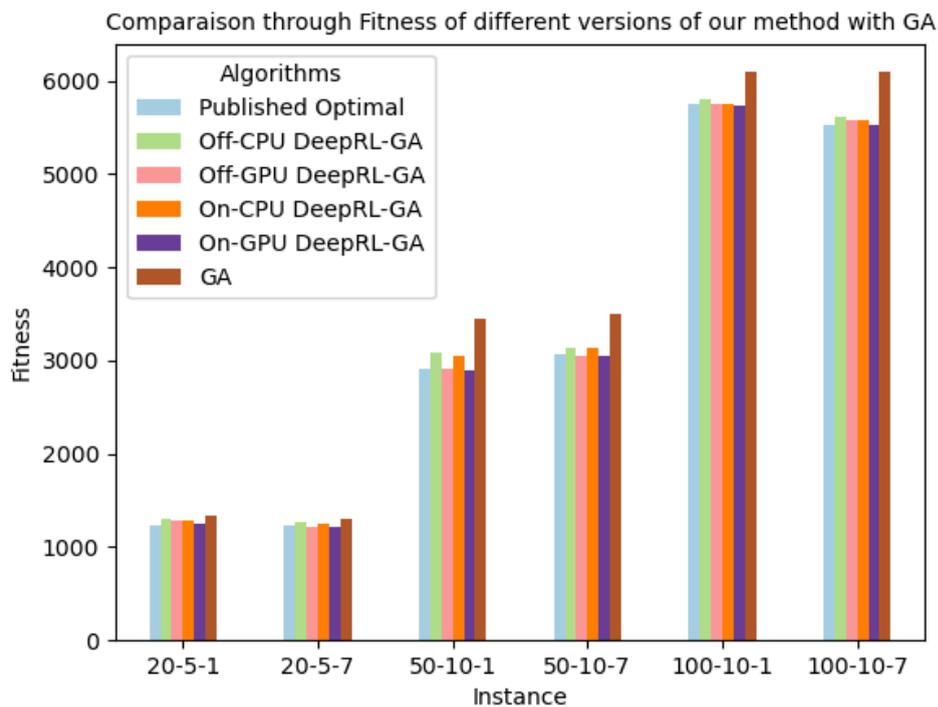

Figure 06. Comparaison through Fitness of different versions of our method with GA

| Instance | NEH | GNEH | CDS | SA | TS | SHC | VNS | NEGA | On-GPU DeepRL-GA |
|---|---|---|---|---|---|---|---|---|---|
| B_20_5_1 | 1334 | 1321 | 1390 | 1297 | 1297 | 1305 | 1297 | 1278 | **1255** |
| B_20_5_7 | 1284 | 1257 | 1393 | 1251 | 1285 | 1251 | 1252 | 1239 | **1214** |
| B_50_10_1 | 3229 | 3249 | 3421 | 3184 | 3156 | 3148 | 3117 | 3071 | **2902** |
| B_50_10_7 | 3271 | 3273 | 3520 | 3247 | 3174 | 3218 | 3165 | 3157 | **3057** |
| B_100_10_1 | 6062 | 5894 | 6209 | 5937 | 5876 | 5925 | 5879 | 5781 | **5731** |
| B_100_10_7 | 5719 | 5710 | 6201 | 5700 | 5690 | 5690 | 5679 | 5641 | **5528** |

Table 06. Computational results of other methods compared with On-GPU DeepRL-GA

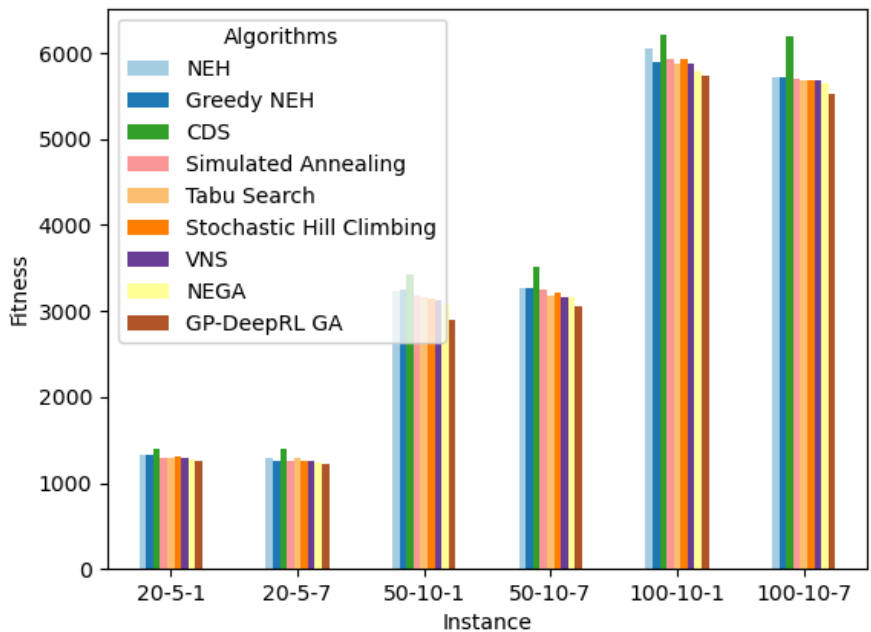

Figure 07. Computational results of other methods compared with On-GPU DeepRL-GA

### 4.2.b. Computation times

Comparaison through time of different versions of our method with GA

| Instance | Off-CPU DeepRL-GA | Off-GPU DeepRL-GA | On-CPU DeepRL-GA | On-GPU DeepRL-GA | GA |
|---|---|---|---|---|---|
| B_20_5_1 | 9.57 | 4.78 | 52.60 | 26.30 | 0.95 |
| B_20_5_7 | 8.28 | 4.12 | 46.56 | 23.28 | 0.98 |
| B_50_10_1 | 20.04 | 10.10 | 279.32 | 139.66 | 18.42 |
| B_50_10_7 | 15.32 | 7.66 | 257.20 | 128.60 | 17.30 |
| B_100_10_1 | 50.67 | 25.32 | 657.63 | 322.09 | 162.94 |
| B_100_10_7 | 48.71 | 24.35 | 536.19 | 305.73 | 161.03 |

Table 07. Comparaison through time of different versions of our method with GA

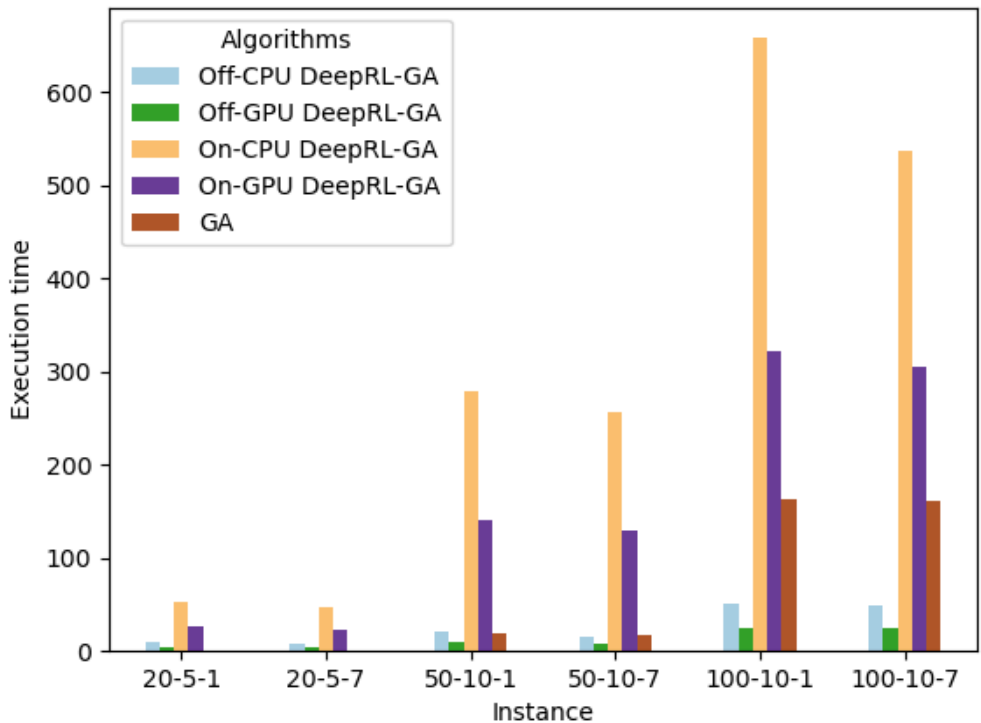

Figure 08. Comparaison through time of different versions of our method with GA

| Instance | NEH | GNEH | CDS | SA | TS | SHC | VNS | NEGA | Off-GPU DRL-GA |
|---|---|---|---|---|---|---|---|---|---|
| B_20_5_1 | 0.02 | 0.33 | 0.00 | 1.97 | 27.07 | 0.09 | 0.13 | 7.04 | **4.78** |
| B_20_5_7 | 0.03 | 1.89 | 0.00 | 1.21 | 27.08 | 0.20 | 0.13 | 9.02 | **4.12** |
| B_50_10_1 | 0.72 | 10.70 | 0.02 | 1.04 | 90.79 | 18.01 | 88.23 | 185.32 | **10.10** |
| B_50_10_7 | 1.73 | 21.66 | 0.05 | 0.94 | 91.68 | 18.47 | 112.96 | 190.59 | **7.66** |
| B_100_10_1 | 6.39 | 157.97 | 0.05 | 2.09 | 319.12 | 93.23 | 225.42 | 1390.71 | **25.32** |
| B_100_10_7 | 12.79 | 296.24 | 0.04 | 2.12 | 288.62 | 99.93 | 181.42 | 1489.98 | **24.35** |

Table 08. Computational results of other methods compared with GPU DeepRL-GA

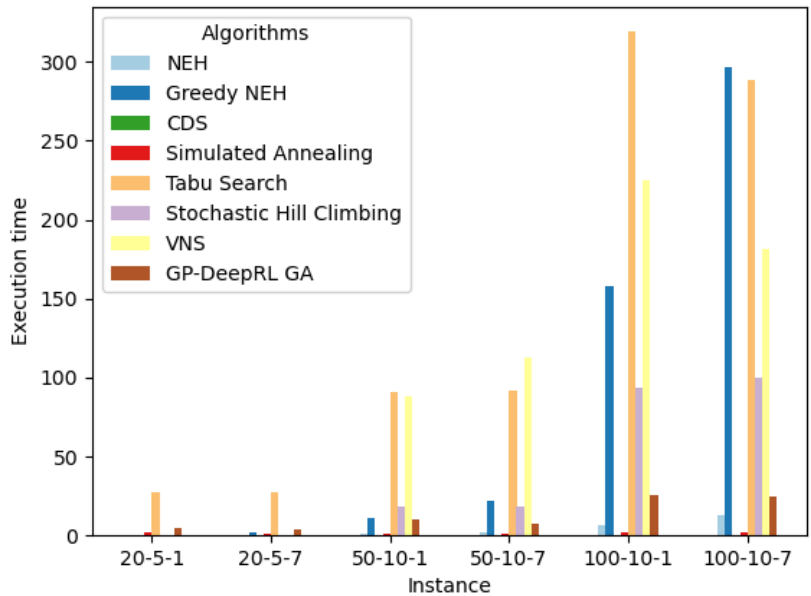

Figure 09. Computational results of other methods compared with GPU DeepRL-GA

### 4.2.c. Experimentation conclusion

Through the computational results, we can notice that our method was effective in finding a good solution, surpassing all of the methods studied through our experimentations, while making a good compromise between the computational time, and the fitness of the solution returned.

## 5. CONCLUSION

This paper introduces a hybrid approach for the Permutation Flowshop Scheduling Problem (PFSP). The method combines three key elements: a Genetic Algorithm (GA), and reinforcement learning Agent, whether it's learning online or offline, and neural network.

To the best of our knowledge, this is the first implementation that integrates Reinforcement Learning into a GA to choose a parent selection operator for PFSP with the objective of minimizing the makespan.

During the execution of the algorithm, all solutions are encoded using a job-based representation (job permutation). The proposed approach, called DeepRL-GA was evaluated on different benchmark instances.

The results demonstrate its efficiency and effectiveness, as it consistently produces competitive solutions within acceptable computational times.

---